\documentclass[letterpaper, 10 pt, conference]{ieeeconf}  

\IEEEoverridecommandlockouts                              

\usepackage{cite}
\usepackage{amsmath,amssymb,amsfonts}
\usepackage{algorithmic}
\usepackage{graphicx}
\usepackage{textcomp}
\usepackage{xcolor}
\usepackage{multirow}
\usepackage{booktabs}
\usepackage[caption=false]{subfig}

\def\BibTeX{{\rm B\kern-.05em{\sc i\kern-.025em b}\kern-.08em
    T\kern-.1667em\lower.7ex\hbox{E}\kern-.125emX}}
    
\title{User Study Exploring the Role of Explanation of Failures by Robots in Human Robot Collaboration Tasks
}

\author{Parag Khanna$^{1}$, Elmira Yadollahi$^{1}$, Mårten Björkman$^{1}$, Iolanda Leite$^{1}$, and Christian Smith$^{1}$
\thanks{$^{1}$ Division of Robotics, Perception and Learning (RPL), EECS, KTH Royal Institute of Technology, Sweden
        {\tt\small paragk@kth.se,}
        {\tt\small elmiry@kth.se,} 
        {\tt\small celle@kth.se,}
        {\tt\small iolanda@kth.se,}
        {\tt\small ccs@kth.se}}%
}

\begin{document}

\maketitle
\thispagestyle{empty}
\pagestyle{empty}

\begin{abstract}
Despite great advances in what robots can do, they still experience failures in human-robot collaborative tasks due to high randomness in unstructured human environments. Moreover, a human's unfamiliarity with a robot and its abilities can cause such failures to repeat. This makes the ability to failure explanation very important for a robot. In this work, we describe a user study that incorporated different robotic failures in a human-robot collaboration (HRC) task aimed at filling a shelf. We included different types of failures and repeated occurrences of such failures in a prolonged interaction between humans and robots. The failure resolution involved human intervention in form of human-robot bidirectional handovers. Through such studies, we aim to test different explanation types and explanation progression in the interaction and record humans.
\end{abstract}

\section{Introduction}
Since the arrival of the fourth industrial revolution, collaborative robots have become very important in the manufacturing and production sectors. They can enhance productivity by working in tandem with humans in a shared workspace with physical interaction \cite{collab-Industry4}. Progressive development in social robots has also paved the way for such robots in households and other social human environments \cite{social_robots_home_n_other}. Inevitably, such robots are prone to failure when they encounter the vast randomness in our environments. We consider the example of pick-and-place tasks, which form a large portion of our common daily activities. Even if we consider a household, the objects to be handled can greatly vary in their shapes, sizes, textures, and other properties. These objects may need to be picked up and placed in congested areas that are difficult to access. A human would easily handle most such objects with high fidelity and precision, which also leads to similar expectations from the collaborative robot. Thus, upon encountering robotic failures, the human potentially loses trust in robots \cite{Correia_fault_justification} and becomes resistant to future interactions with them. This forms the motivation to incorporate explanations of failures in collaborative robots for HRC and invoke human help for failure resolution \cite{das2021explainable}. 

With the user study presented in this work, we aimed to explore the effect of different levels of explanation on human perception and trust. In the prolonged interaction, we also aimed to compare constant levels with decaying levels of explanation.
\section{Prior work}
In \cite{Correia_fault_justification}, the impact of a collaborative robot's failure on human trust as well as the impact of justification strategies were explored in a user study. Their findings show that an incorrect robot is regarded as far less trustworthy. They also show that justifications were able to reduce the impact of failures on human trust for failures with minor consequences.
It was aimed at autonomously generating explanations for unexpected failures in a pick-and-place task for a robot in a home environment in \cite{das2021explainable}. The explanation was especially focused on obtaining failure recovery with the assistance of non-expert users in the HRC. It was observed that failure and solution identification are most successfully achieved by explanations that include the context of the failure action and the history of the prior actions.
\begin{figure}[t]
      \centering
        \includegraphics[width=.99\linewidth,height=6cm,trim={4.1cm 1.0cm 4.1cm 0.0cm},clip]{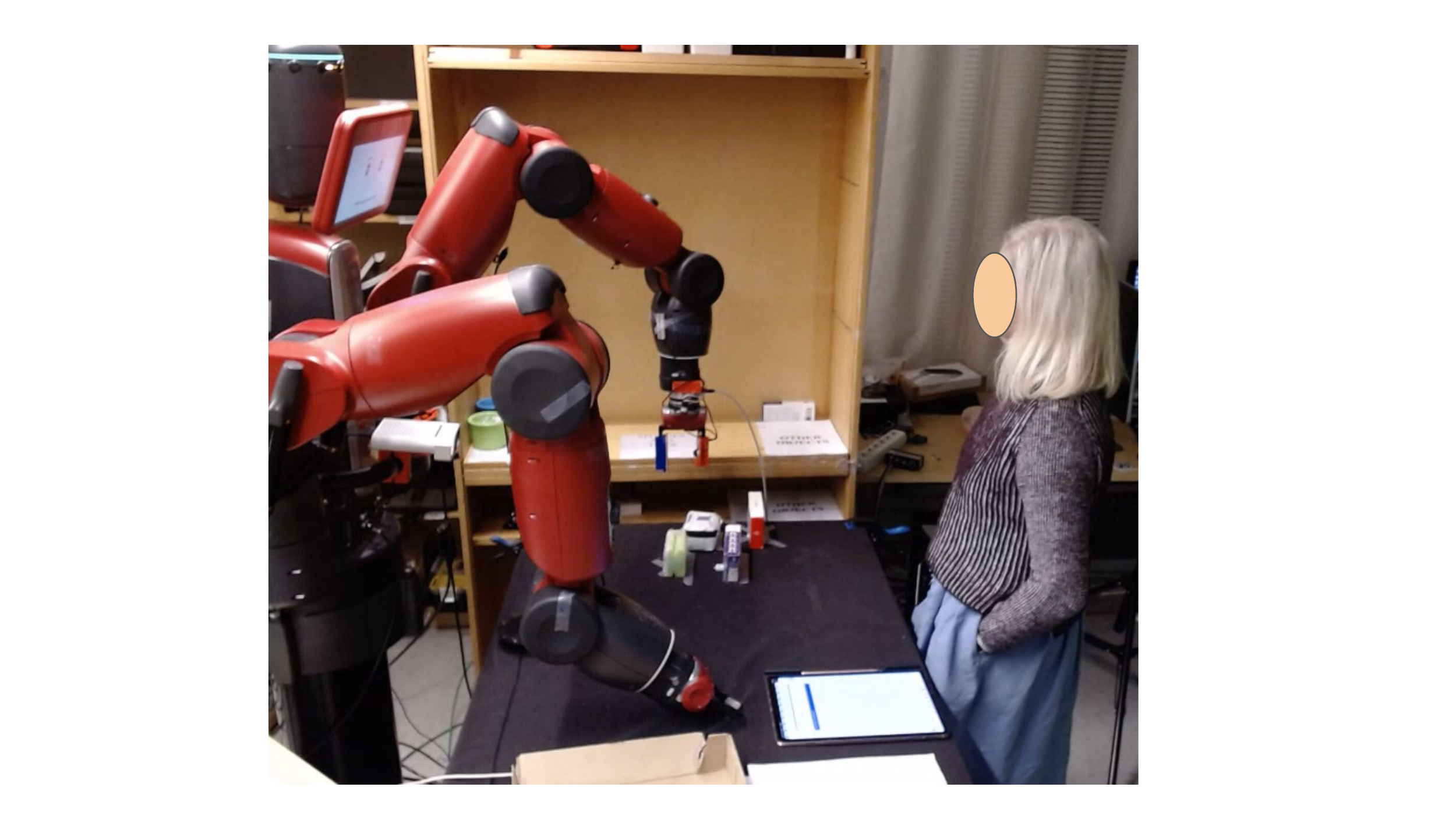}  
     \setlength\abovecaptionskip{-0.2\baselineskip}
    \caption{The Human-Robot Collaboration scenario: Human is tasked to place objects in front of the robot at the marked locations and the Robot is tasked to pick the objects from the table and place them on the shelf}
      \label{fig:HRC_baxter}
\end{figure}
\section{Study with Failures and Explanation}
\subsection{Motivation}
In the user study, we aimed to create a human-robot collaborative interaction with robotic failures. To complete an assigned task, robots generally need to complete a series of sub-tasks. In a collaborative setup, the robot and the human work together, performing specific sub-tasks and complementing each other. However, if the robot fails to carry out a sub-task and it hinders the accomplishment of the final goal of task completion, the human could lose trust in the robot for further interaction. For a naive human user, unfamiliarity with the robot's abilities could also lead to such failures. As seen in the literature, one possible method for the robot to regain trust is by providing an explanation of its failure \cite{Correia_fault_justification}. Since it is a collaborative setup, the robot can also ask for human assistance by stating the necessary failure resolution for the sub-task completion \cite{das2021explainable}. By getting proper human help, the collaborative task can be jointly completed despite the failure.

This motivated us to create a user study with a common goal for the robot and the human, where the robot has to complete some sub-tasks to achieve the goal. We further included failures for these sub-tasks with failure explanation and resolution to be given by the robot. In this study, we aimed to see the effect of different types of explanations on human trust and perception of the robot, along with the human-robot joint performance in the task.
\subsection{Study Design}
For the HRC task, a Baxter robot was given the task of stocking a shelf by placing objects on it. The human would deliver these objects to the robot by placing them on the table in front of them. To get a prolonged interaction, we incorporated four rounds of four objects each to be placed on the shelf. The objects, marked with letters A,  B,  C, or D, were to be taken out of a big box by the human and placed at the corresponding location marked by the same letters on the table at the start of a round. 
The robot must successfully complete the following actions while handling an object: identify it, pick it up, carry it, and ultimately place it on the shelf. We examine probable robotic failures in these sub-steps in our study design. When the robot encounters a failure, it needs human assistance to accomplish the task. The following failures and necessary resolutions were devised and incorporated:
\begin{figure}[ht]
      \centering
        \includegraphics[width=.8\linewidth,height=5cm,trim={0.1cm 0.0cm 0.1cm 0.0cm},clip]{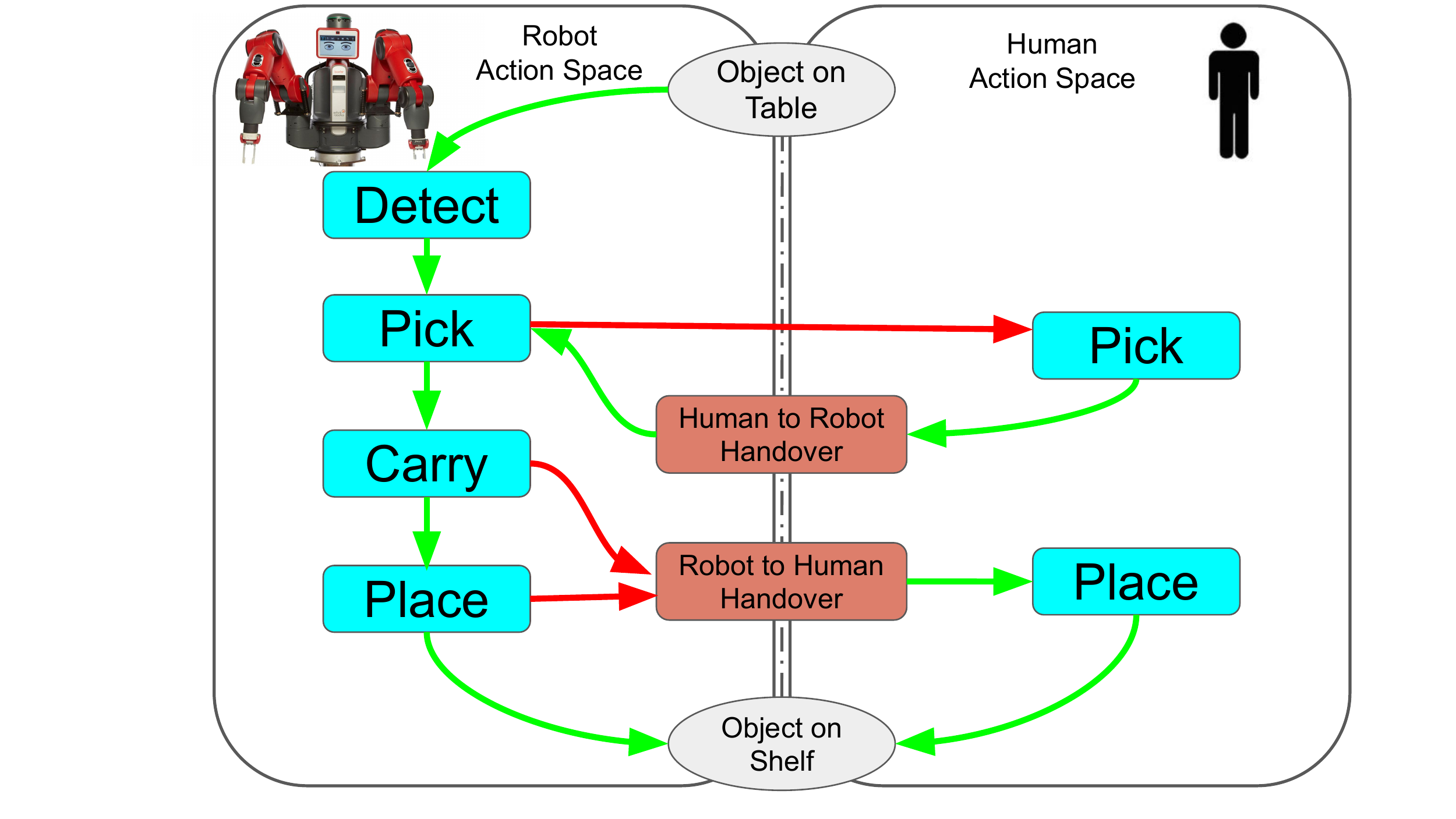}  
     \setlength\abovecaptionskip{-0.2\baselineskip}
    \caption{Description of human-robot collaborative task with the robot and human action spaces. Arrows in green represent transitions due to action success. Arrow in red represents transitions due to action failure. }
      \label{fig:HRC_collab_task}
\end{figure}
\begin{enumerate}
    \item \textbf{Pick Failure}: In the way, it is placed on the table, the object does not fit in the gripper of the robot as it tries to pick it. \\
    \textbf{Resolution}: Human picks the object and places the object in the robot gripper, i.e., a human-to-robot handover. 
    \item \textbf{Carry Failure }: The object is too heavy for the robot arm to safely carry. \\
    \textbf{Resolution}: Human takes the object back from the robot by a robot-to-human handover. Human places the object on the shelf.
    \item \textbf{Place Failure}: Robot can not place the object as it failed to reach the lower portion of the shelf which is the desired location for the object. \\
    \textbf{Resolution}: Human takes the object back from the robot by a robot-to-human handover. Humans place the object in the shelf.
\end{enumerate}

\begin{table}[b]
\caption{Round-wise Description}
\begin{center}
\begin{tabular}{c c c c c}
     \toprule
     
 Round & Object & Object Type & Failure \\ [0.5ex] 
    \midrule
    \multirow{4}{*}{Round 1}& A & Sponge &  None \\
    & B & Cloth Bag & F1-Pick \& F2-Carry  \\
    & C & Random Box1 & None\\
    & D & Pen Box &F3-Place  \\  
    \midrule 
    \multirow{4}{*}{Round 2}& A &Heavy Box 1 &F2-Carry \\
    & B &Random Box 2& None \\
    & C &Flat Box& F1-Pick \\
    & D &Toy& None  \\
    \midrule
    \multirow{4}{*}{Round 3}& A &Pen Box& F3-Place \\
    & B &Pen Box Heavy& F2-Carry  \\
    & C &Toy& None  \\
    & D &Random Box 3& None \\ 
    \midrule
    \multirow{4}{*}{Round 4}& A&Flat Box & F1-Pick  \\
    & B &Sponge& None \\
    & C &Pen Box& F3-Place \\
    & D &Heavy Box 2& F1-Pick \& F2-Carry \\ 
    \bottomrule
\end{tabular}
\end{center}
\label{tab:HRC_round_objects}
\end{table}

\begin{figure*}[t]
\captionsetup[subfigure]{justification=centering}
\centering
 \subfloat[Sponge and Toy]{
\includegraphics[width=.18\linewidth]{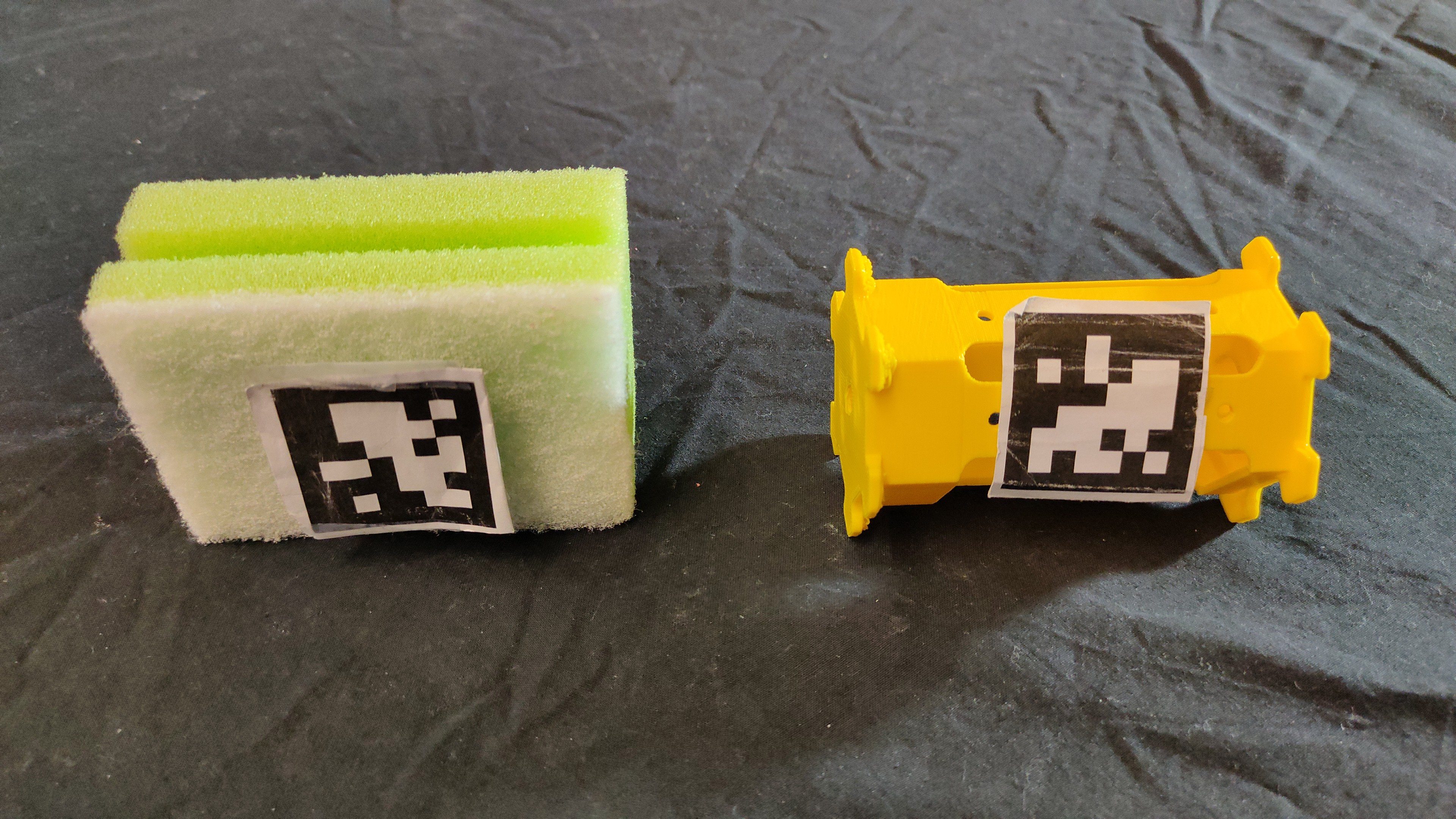}
}
\subfloat[Pen box (left) and Pen box-heavy (right)][Pen box (left), \\ Pen box-heavy (right)]{
\includegraphics[width=.18\linewidth]{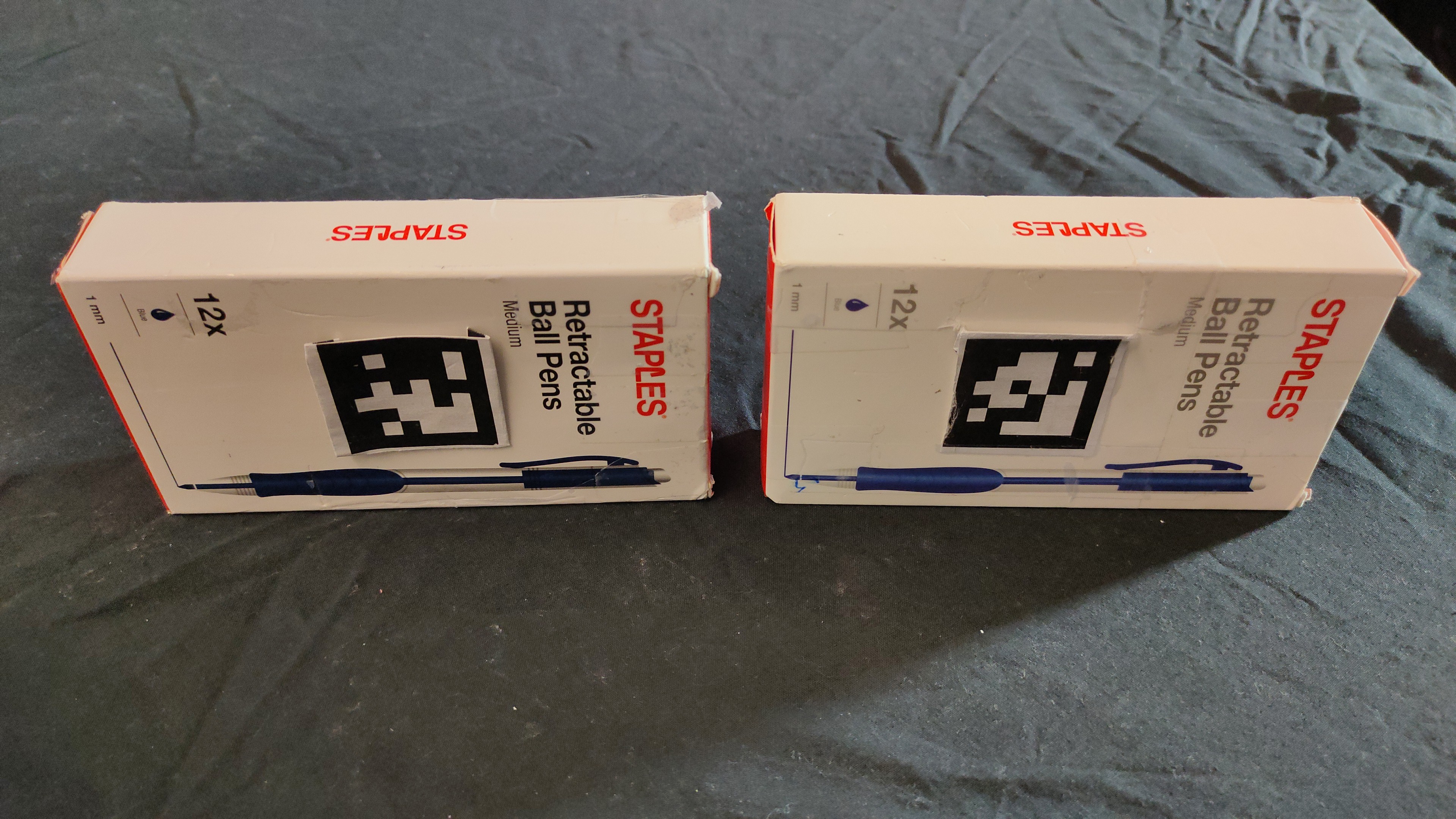}
}
\subfloat[Heavy box 1 (left), Heavy box 2 (right)][Heavy box 1 (left), \\ Heavy box 2 (right)]{
\includegraphics[width=.18\linewidth]{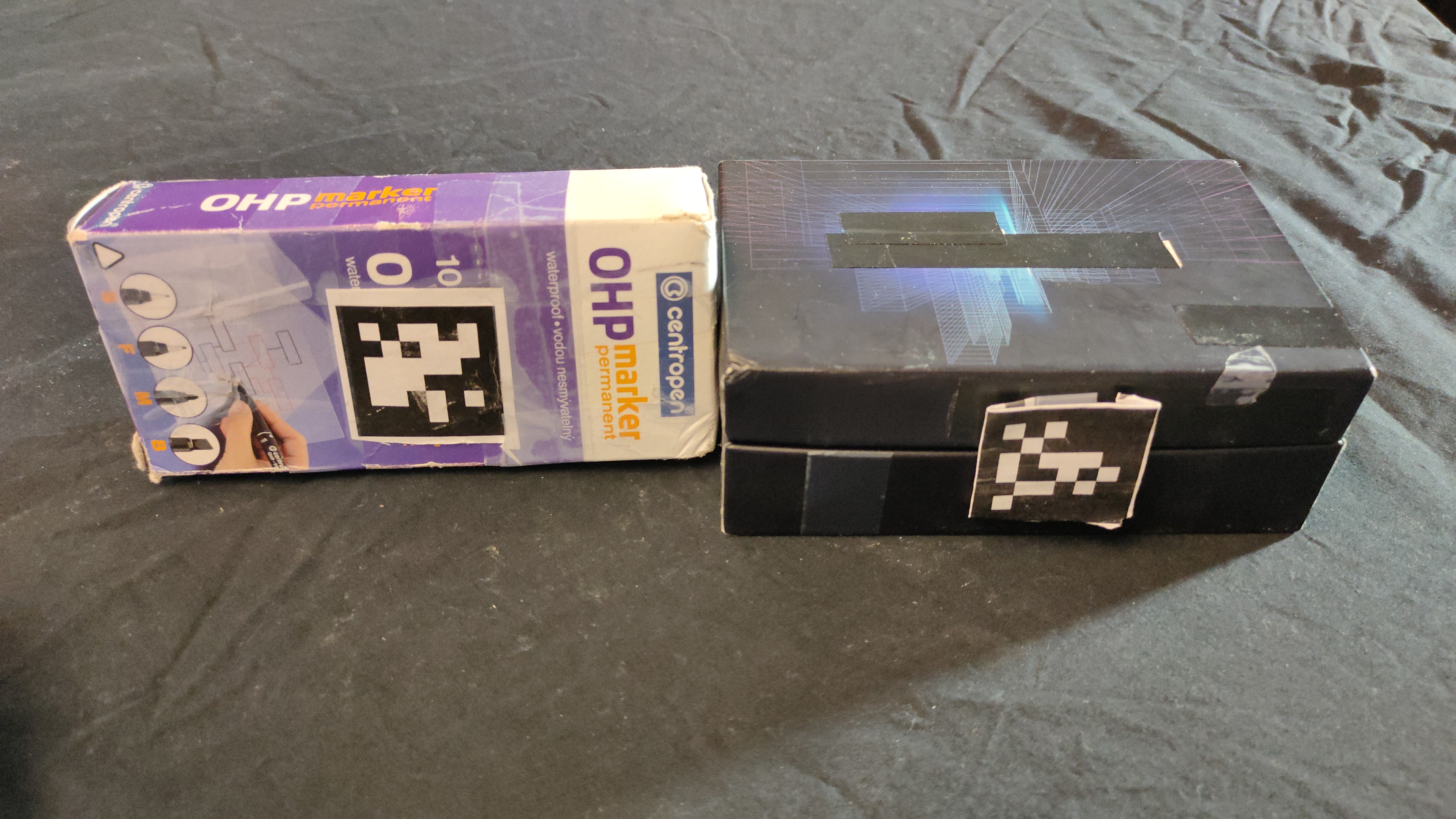}
}
\subfloat[Cloth box, Flat box][Cloth box, \\ Flat box]{
\includegraphics[width=.18\linewidth]{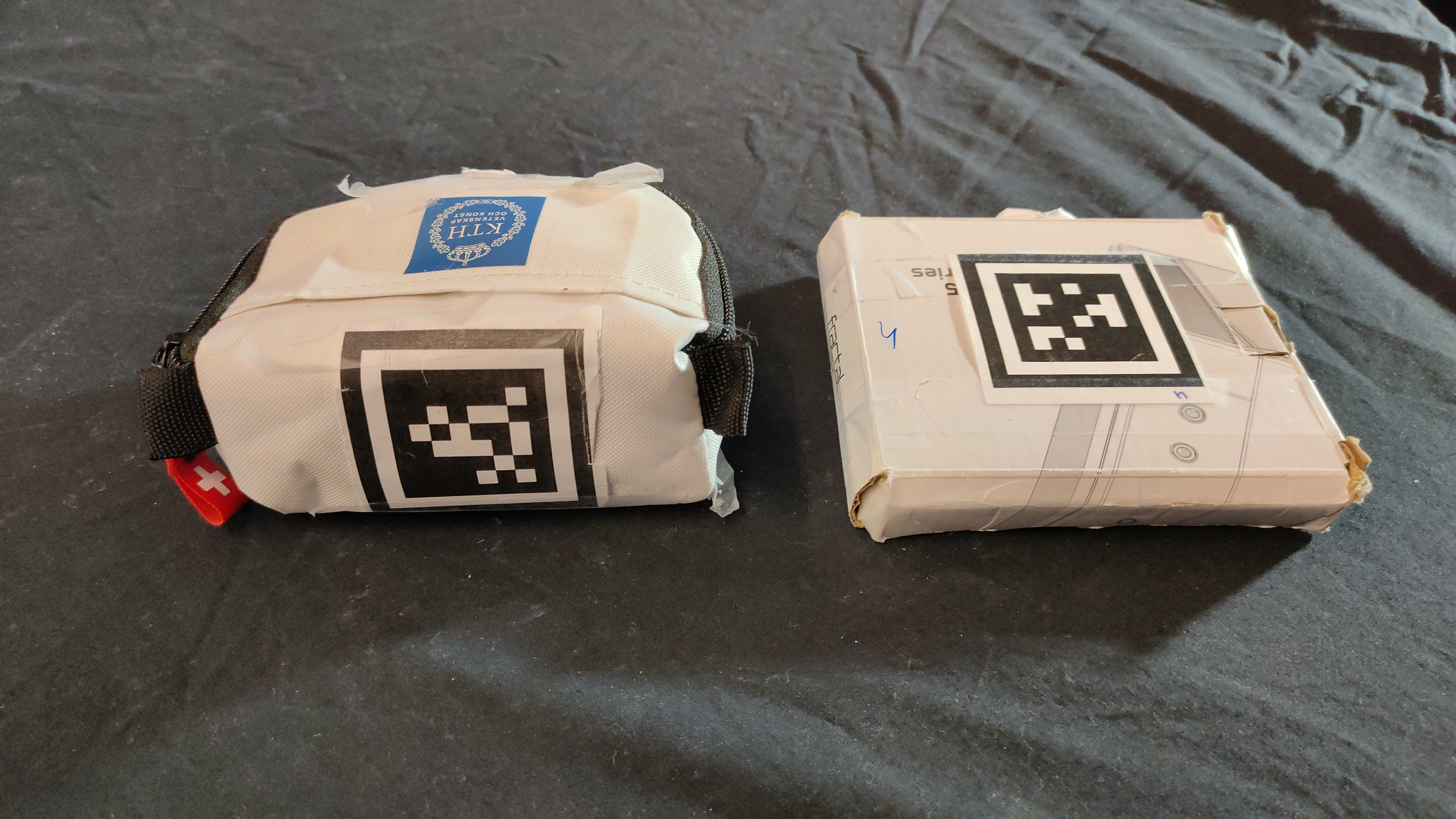}
}
\subfloat[Random Box 2 (left), Random Box 1 (right)][Random Box 1,3 (left)\\ Random Box 2 (right)]{
\includegraphics[width=.18\linewidth]{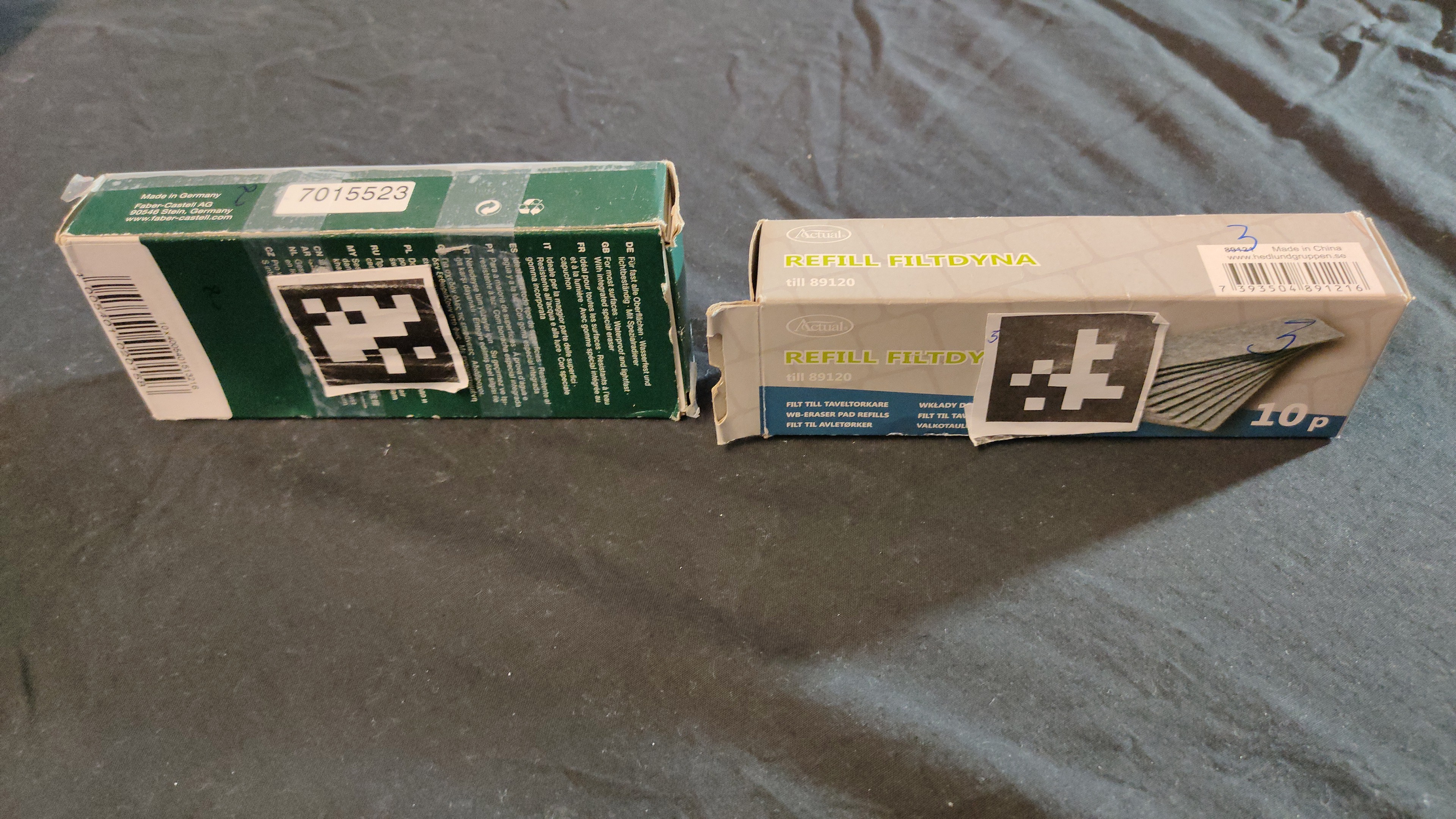}
}
\caption{Objects}
\label{fig:Objects}
\end{figure*}

The potential process for moving an object from the table to the shelf is depicted in Fig \ref{fig:HRC_collab_task}. Failures of robotic action are indicated in red and require human action as a resolution. As indicated in Table \ref{tab:HRC_round_objects}, across four rounds, we evenly distributed the seven objects for which the robot completes all actions and the nine objects for which there were some robotic failures. In particular, the pen boxes go to the lower part of the shelf, which is not reachable by the robotic arm.
The objects chosen are items commonly used at home and at work, as illustrated in Fig. \ref{fig:Objects}. We also wanted to make sure that when a person sees and places a specific object on a table, the accompanying robotic failure is not immediately apparent to them. The robotic failures are pre-programmed with specific actions to be performed by the robot to convey those failures.
\subsection{Explanation Levels}
We considered four levels of explanations based on \cite{das2021explainable}. The failure resolution statement was also adapted to the explanation.
\begin{enumerate}
    \item Level 0 - Non-verbal explanation: This included a robot head shake at each failure and the failure action itself. The robot arm attempted to pick up each object, and if a pick failed, it shook its head and went into the handover pose. In the event of a carry object failure, the robot lifted the arm with the object in the gripper and imitated a falling down motion of the arm accompanied by a head shake. For place failure, the robot arm tried to go below the upper level of the shelf and the table but failed, and a head shake occurred. Following both carry and place failures, the robot arm then went into a handover pose and asked the human to take the object back. 
    Thus, a level 0 explanation is the minimum explanation that always followed a failure.
    \item Level 1 - Action based: After the failure, the robot stated the failure action and the resolution action.
    \begin{itemize}
        \item "I failed to pick up the object", "Hand it to me".
        \item "I failed to carry the object", "Carry it for me".
        \item "I failed to pick up the object", "Place it for me".        
    \end{itemize}
    \item Level 2 - Context-based: After the failure, the robot told the failure and the cause of failure, followed by the resolution statement.
    \begin{itemize}
        \item "I can't pick up the object because it doesn't fit in my gripper", "Can you hand it over to me".
        \item "I can't carry the object because it is too heavy for my arm.", "Can you carry the object for me"
        \item "I can't place the object because the destination is out of my arm's reach, Can you place the object for me".        
    \end{itemize}
    
    \item Level 3 - Context + History based: After the failure, the robot stated the previous successfully completed action with the current failure action and its cause. The resolution statement also further explains how to do the resolution action.
    \begin{itemize}
        \item "I can detect the object, but I can't pick it up because it doesn't fit in my gripper.", "Can you hand it over to me by placing it in my gripper?".
        \item "I can pick up the object, but I can't carry it because it is too heavy for my arm.", "Can you carry the object for me and place it on the shelf?"
        \item "I can carry the object, but I can't place it because the destination is out of my arm's reach.", "Can you place the object on the shelf location that is out of my reach?".        
    \end{itemize}
\end{enumerate}
\subsection{Explanation Progression}
We chose a four-round HRC task with repeated failures to study the effects of failure-explanation in the long-term interaction. In the four rounds, keeping a constant level of explanation helped us measure the human perception of the robot when the human is exposed to a given fixed level of failure explanation for repeated failures. We also believe it can be interesting to see the effects of decreasing the level of explanation when the failures repeat. We wanted to test if a decay in the explanation level could lead to better or similar results than a fixed high level of explanation.
Thus, we had two types of progressions where the explanation levels remained fixed or decayed:
\begin{itemize}
    \item Fixed Level of explanation: Across four rounds, the level of explanation is kept constant. 
    \item Progressive decay in explanation: The level of explanation started from a maximum of level 3 in round 1 and decayed like level 2 in round 2, level 1 in round 3, and non-verbal level 0 in round 4. 
    \item One step drop-in explanation: The level of explanation started from a maximum of level 3 in round 1 and remained constant at level 0 or level 1 for rounds 2,3 and 4. We included this type of decay to see the effects when robot provides a high level of explanation for a failure once and a lower level of explanation if the failure occurs again.

\end{itemize}

\subsection{Procedure}
Before the start of the study, they were introduced to the robot and its goal of filling the shelf with objects. They were not made aware of the possibility of robotic failure during the task. Once the study started, the participant would only interact with the robot.

The interaction started with the robot welcoming and asking the human to place a new set of objects on the table. 
The robotic arms are in the home position as seen in Fig. \ref{fig:HRC_baxter} and only the left arm is used in the experiments. 
The robot waits and asks if the person has placed the objects on the table. 
Once the person places the objects, the robot starts the task and for each object, the robot follows the following steps:
\begin{enumerate}[] 
     \item \textbf{Detect object:} Robot announces the object it is going for and scans the table for the corresponding object. 
     If it succeeds, it moves to step 3, else it proceeds to step 2.
     \item \textbf{Failure $f_0$ and $r_0$:} Robot provides the detection failure explanation and repeats step 1 until moving to step 3. 
     \item \textbf{Pick object:} Robot starts by going over the object and lowering its gripper around it to pick it up. 
     If it succeeds in picking up, it goes to step 5, else proceeds to step 4. 
     \item \textbf{Failure $f_1$ and $r_1$:} Robot explains the failure and goes to the handover pose to receive the object from the human. 
    It always asks the human to confirm if it can close its gripper by asking ``Should I close the gripper?'' and once it closes the gripper it moves to step 5. 
    \item \textbf{Carry object:} Robot tries to lift the object up by showing an upward motion of the arm. 
     If the robot successfully carries the object, it moves to step 8, else if any failure occurs, it moves to step 7. 
     \item \textbf{Failure $f_2$:} For carry failure, the robot shows a controlled falling-down motion with its arm and explains the failure followed by going to the handover pose. 
     Then, it asks the human to take the object back and proceeds to step 9. 
     \item \textbf{Place Object:} Robot carries the object to the shelf from the table. 
     In the case of objects with place failure in the lower shelf, it moves to step 10. 
     For objects with no intended failure, it places them at a predefined location on the upper shelf, the arm returns to its home position and the robot moves on to the next object. 
     \item \textbf{Failure $f_3$:} For place failure, the object should be placed on the lower shelf that is not accessible to the robot arm. 
     For these objects, the robot executes a predefined trajectory where the arm goes down towards the lower shelf but it fails to place the object.
     Then, it provides the explanation and moves to the handover pose, asking the human to take the object back and proceed to step 9.    
     \item \textbf{Robot to human handover and $r_2$/$r_3$:} The robot-to-human handover happens automatically with the gripper opening when a sufficient pull force of 3N is detected, aimed at the least likelihood of the object dropping, based on a prior study \cite{parag-humanoids}. 
     If the human fails to take the object, the gripper opens on its own after a fixed time interval. Once the human takes the object, the robot states the resolution needed, expecting the human to place the object on the shelf. 
 \end{enumerate}
 The robot repeats the resolution statement if the human does not finish the resolution activity in a specific amount of time or if the human asks a question. 
In this experiment, a Wizard of Oz paradigm governs this repetition. The robot repeats the resolution a maximum of five times, each spaced by three seconds. After 5 repetitions, the robot then moves on to the next object by announcing the next object.

Once the last object is handled for a round, the participant would rank the interaction by answering some questions on a nearby tablet regarding the interaction. The robot asked for a new set of objects once the participant was done answering the questions. At the end of round 4, the participant would also answer a detailed set of questionnaires regarding the whole experiment.

Aside from subjective responses, we also recorded participant behavior in terms of audio-visual data and upper body tracking with a depth-vision sensor. 

\section{Conclusion and Future Work}
We present a user study to examine the effects of different levels of explanations of failures for HRC tasks. For a robotic pick and place task, we discussed the possible failures incorporated in the study and their possible resolutions via human-robot handovers. We described different levels of explanation for failures considered in our experiments. To measure the effects of a prolonged interaction, we proposed and applied different explanation progression strategies. 

For future work, we would evaluate and compare different explanation levels against each other based on subjective parameters like human perception and trust. We would also see the impact of the explanation on the resolution action completion by the participants, which shows human understanding of the failure and its resolution.








\bibliographystyle{IEEEtran}
\bibliography{biblio}

\end{document}